\documentclass[runningheads]{llncs}
\usepackage{graphicx}      
\usepackage{amsmath,amssymb,amsfonts}
\usepackage{bm, algorithmic}
\usepackage{textcomp}
\usepackage{xcolor}
\usepackage{hyperref}
\usepackage{subfiles}

\begin{document}


\title{Towards Stochastic Fault-tolerant Control using Precision Learning and Active Inference}

\titlerunning{Towards Stochastic FT Control.}

\author{Mohamed Baioumy$^1$ \and Corrado Pezzato$^2$ \and Carlos Hernández Corbato$^2$\and Nick Hawes$^1$ \and Riccardo Ferrari$^3$}


\authorrunning{M. Baioumy et al.}

%


\institute{Oxford Robotics Institute, University of Oxford\\
\email{\{mohamed, nickh\}@robots.ox.ac.uk}
\and Cognitive Robotics, Delft University of Technology\\ \and Delft Center for Systems and Control, Delft University of Technology\\
\email{\{c.pezzato, c.h.corbato, r.ferrari, m.wisse\}@tudelft.nl}\\}

\maketitle              
%
\begin{abstract}

This work presents a fault-tolerant control scheme for sensory faults in robotic manipulators based on active inference. In the majority of existing schemes a binary decision of whether a sensor is healthy (functional) or faulty is made based on measured data. The decision boundary is called a threshold and it is usually deterministic. Following a faulty decision, fault recovery is obtained by excluding the malfunctioning sensor.
We propose a stochastic fault-tolerant scheme based on active inference and precision learning which does not require a priori threshold definitions to trigger fault recovery. Instead, the sensor precision, which represents its health status, is learned online in a model-free way allowing the system to gradually, and not abruptly exclude a failing unit.
Experiments on a robotic manipulator show promising results and directions for future work are discussed.
\section{Introduction}


Safety is paramount for autonomous systems designed for operating in the real world. External dangers in the environment such as steep and slippery terrain encountered by planetary rovers \cite{slip_planetary_rover} can compromise entire missions. In addition to external dangers, internal system components can also fail and possibly lead to dangerous outcomes if a proper fault-tolerant (FT) control scheme is not present. Building systems that are robust to the presence of faulty components, such as sensors and actuators, is addressed in the FT literature \cite{sensorFault1,sensorFault2,ChengBook}. Generally speaking, FT control consists of \textit{fault detection}, which provides a signal representing whether a system component is faulty; \textit{fault isolation}, which identifies the exact faulty component, and \textit{fault recovery}, which typically contains a switching or a re-tuning procedure of the running controllers to accommodate for the fault. 

Several methods are available for fault detection, but model-based methods are among the most powerful and appealing, as they provide theoretical guarantees \cite{ChengBook}. These methods rely on monitoring system outputs using mathematical models to generate `symptoms' called \emph{residual signals}. These signals are then compared to carefully designed detection \emph{thresholds}: the sensor is `faulty' if a threshold is exceeded or `healthy' otherwise. To recover from a fault, the recovery actions are usually performed through controller reconfiguration, that entails adapting the controller parameters, or switching to another controller or to backup sensors and actuators \cite{Narendra}. 
When modelling external dangers or monitoring faulty systems, robust detection thresholds are essential. Robust thresholds used in existing work (such as \cite{budd2020safeest} or \cite{sensorFault2}) are often \textit{deterministic}, but this is sub-optimal. For instance, if the safety threshold for a rover on a slippery terrain slope is 15 degrees, this means that a slope of 14.9 is safe but 15.1 is unsafe. Additionally, a slope of 15.1 degrees and 40 degrees are `equally unsafe'. 

In this paper we build upon two ideas in the literature. First, the usage of a stochastic fault tolerant formulation (e.g. \cite{fang2015stochastic-ft,rostampour2020privatized}). This allows the agent to overcome the issues mentioned above. Additionally, we leverage an unbiased active inference controller (u-AIC) \cite{baioumy2021ECC}, evolved from previous active inference controllers (AIC) \cite{buckley,baioumy2020active,Pezzato2020}. Active inference is a promising framework for FT control which has already been shown to facilitate fault-detection, isolation and recovery for robotic systems with sensory faults \cite{baioumy2021ECC,Corrado2020iwai}. 

Besides fault tolerance, active inference showed promising performance in many control and state-estimation problems in robotics \cite{lanillos1,lanillos2}. Particularly interesting are the works on robot arm control \cite{Pezzato2020,oliver,lopez}, which highlighted the adaptive properties of active inference. Active inference also shares similarities with the control as inference framework \cite{levine2018reinforcement}.  A more extensive analysis of active inference and its relation to control as inference can be found in \cite{millidge2020relationship,imohiosen2020active}.

The main contribution of this paper is a FT controller for robot manipulators with sensory faults based on unbiased active inference with a stochastic decision boundary. Unlike previous work \cite{baioumy2021ECC}, here we model the precision (inverse covariance) of each sensor in our system and determine the probability of the sensor being healthy to be proportional to its precision. Our approach allows for fault-tolerant behaviour without needing any threshold definition a priori, and without the need to design additional ad-hoc recovery mechanisms. Finally, this work can be used stand-alone or in conjunction with other methods for fault-detection and isolation in order to estimate the faults.

\section{Problem statement and background}

The FT scheme in this paper is derived for a class of systems, namely serial robot manipulators equipped with sensors for joint position and velocity, and end-effector location. In the following, the problem and the setup are described, and some background knowledge on u-AIC for torque control from \cite{baioumy2021ECC} is presented. 

\subsubsection{Problem Setup.}
Consider a robotic manipulator with state $\bm x$ comprising of its joint positions and velocities $\bm{x} = [q \,\, \dot{q}]^\top$. The available sensors provide noisy joint position and velocities $\bm y_q,\ \bm y_{\dot{q}}$ readings. In addition, the end-effector Cartesian position $\bm y_v$ is available through a visual sensor. The system's output is represented by $\bm y = [\bm y_q,\ \bm y_{\dot{q}},\ \bm y_v]\in\mathbb{R}^d $. The proprioceptive sensors and the visual sensor are affected by zero mean Gaussian noise $\bm \eta=[\bm \eta_q$, $\bm \eta_{\dot{q}}$, $\bm \eta_v]$. Additionally, the visual sensor is affected by barrel distortion. The system is controlled through an u-AIC \cite{baioumy2021ECC} which steers the robot arm to a (changing) desired configuration in joint space $\bm \mu_d$, providing the control input $\bm u\in\mathbb{R}^m$ as torques to the joints.

\subsubsection{Background: Unbiased Active Inference controller}
\label{subsec:unbiased_aic}

In this section we briefly describe the  u-AIC as introduced in \cite{baioumy2021ECC}, to which an interested reader is referred for more details on the derivations of the following equations. The novel FT method presented in this paper in Sec.~\ref{sec:precision_learning_gradient_on_f} builds upon the u-AIC, but instead of employing an ad-hoc hard update of the precision of a faulty sensor after fault detection, it relies on online precision learning during operations.

Let us consider $\bm{x} = [q \,\, \dot{q}]^\top$ and let us define a probabilistic model where actions are modelled explicitly:
\begin{equation}
    p(\bm{x}, \bm{u}, \bm y_v, \bm y_q, \bm y_{\dot{q}}) = \underbrace{p(\bm{u}|\bm{x})}_{control} \underbrace{p(\bm y_v|\bm{x}) p(\bm y_q|\bm{x}) p(\bm y_{\dot{q}}|\bm{x})}_{observation \hspace{1 mm}model} \underbrace{p(\bm{x})}_{prior} 
    \label{eq: factorization_u-AIC_joint_probabilistic_distribution_ with_explicit_actions}
\end{equation}
Note that with the u-AIC the information about the desired goal to be reached is encoded in the distribution $p(\bm u | \bm x)$. In this paper, as in \cite{baioumy2020active}, we assume that an accurate dynamic model of the system is not available to keep the solution system agnostic and to highlight once again the adaptability of the controller.

The u-AIC aims at finding the posterior over states as well as the posterior over actions $p(\bm{x}, \bm{u}| \bm y_v, \bm y_q)$. The posteriors are approximated using a variational distribution $Q(\bm{x}, \bm{u})$. We can make use of the mean-field assumption ($Q(\bm{x}, \bm{u}) = Q(\bm{x})Q(\bm{u})$) and the Laplace approximation, and assume the posterior over the state $\bm{x}$ Gaussian with mean $\bm{{\mu}}_{x}$ \cite{variational}. Similarly for the actions, the posterior $\bm{u}$ is assumed Gaussian with mean $\bm{\mu}_{u}$. By defining the Kullback-Leibler divergence between the variational distribution and the true posterior, one can derive an expression for the so-called free-energy $F$ as \cite{baioumy2021ECC}:
\begin{equation}
     F = -\ln p(\bm{\mu}_{u}, \bm{{\mu}}_{x}, \bm y_v, \bm y_q, \bm y_{\dot{q}}) + C
\end{equation}
Considering eq.~\eqref{eq: factorization_u-AIC_joint_probabilistic_distribution_ with_explicit_actions} and assuming Gaussian distributions, $F$ becomes:
\begin{equation}
\begin{split}
    \label{eq:laplace_F_final_vector}
    F  
    &=  \frac{1}{2}(\bm{\varepsilon}_{y_q}^\top \Sigma_{y_q}^{-1}\bm{\varepsilon}_{y_q}
    +  \bm{\varepsilon}_{y_{\dot{q}}}^\top \Sigma_{y_{\dot{q}}}^{-1}\bm{\varepsilon}_{y_{\dot{q}}}
    + \bm{\varepsilon}_{y_v}^\top \Sigma_{y_v}^{-1}\bm{\varepsilon}_{y_v}\\
    &+ \bm{\varepsilon}_{x}^\top \Sigma_{x}^{-1}\bm{\varepsilon}_{x}
    + \bm{\varepsilon}_{u}^\top \Sigma_{u}^{-1}\bm{\varepsilon}_{u} + \ln|\Sigma_{u}\Sigma_{y_q}\Sigma_{y_{\dot{q}}}\Sigma_{y_v}\Sigma_{x}|)
    + C,
\end{split}
\end{equation}

The terms $\bm{\varepsilon}_{y_q} = \bm{y}_q-\bm{\mu}$, $\bm{\varepsilon}_{y_{\dot{q}}}=\bm y_{\dot{q}}-\bm{\mu}'$, $\bm{\varepsilon}_{y_v}=\bm{y}_v-\bm{g_{v}}(\bm{\mu})$ are the sensory prediction errors respectively for position, velocity, and visual sensory inputs. The controller represents the states internally as $\bm \mu_x = [\bm{\mu},\ \bm{\mu}']^\top$. The relation between internal state and observation is expressed through the generative model of the sensory input $\bm g = [\bm g_q,\ \bm g_{\dot{q}},\ \bm g_v]$. Position and velocity encoders directly measure the state, thus $\bm g_q$ and $\bm g_{\dot{q}}$ are linear (identity) mappings. To define $\bm g_v$, instead, we use a \emph{Gaussian Process Regression} (GPR). This is particularly useful because we can model the noisy and distorted sensory input from the camera, and at the same time we can compute a closed form for the derivative of the process with respect to the beliefs $\bm \mu$, required for the state update laws. For details, see \cite{baioumy2021ECC}.

Additionally, $\bm{\varepsilon}_{u}$ is the prediction error on the control action while $\bm{\varepsilon}_{x}$ is the prediction error on the state. \textcolor{black}{The latter is computed considering a prediction of the state $\hat{\bm x}$ at the current time-step such that $ \bm{\varepsilon}_{x} = (\bm{\mu}_x - \hat{\bm x} )$. The prediction is a deterministic value $\hat{\bm x} = [\hat q \,\, \hat{ \dot{q}}]^\top$ } which can be computed in the same fashion as the prediction step of, for instance, a Kalman filter. The prediction is approximated propagating forward in time the current state belief using the following simplified discrete time model:
\begin{equation}
    \label{eq:euler_integration_a}
    \hat x_{k+1} = 
    \begin{bmatrix}
    I & I \Delta t \\
    0 & I
    \end{bmatrix}
    \mu_{x,k}
\end{equation}
where $I$ represents an unitary matrix of suitable size.
This form assumes that the position of each joint is thus computed as the discrete time integral of the velocity, using a first-order Euler scheme. This approximation can be avoided if a better dynamic model of the system is available, and in that case predictions can be made using the model itself. Finally, by choosing the distribution $p(\bm u | \bm x)$ to be Gaussian with mean $f^*(\bm{\mu}_x, \bm{\mu}_d)$, we can steer the systems toward the target $\bm{\mu}_d$ without biasing the state estimation. This results in  $\bm{\varepsilon}_{u} = (\bm{\mu}_u - f^*(\bm{\mu}_x, \bm{\mu}_d))$. 

In the u-AIC state-estimation and control are achieved using gradient descent the free-energy. This leads to:
\begin{equation}
    \label{eq:u-AIC_F_minimize_a}
    \dot{\bm{\mu}}_u = - \kappa_{u}\frac{\partial F}{\partial \bm{\mu_u}}, \hspace{3 mm}
    \dot{\bm{{\mu}}}_x = - \kappa_{\mu}\frac{\partial F}{\partial {\bm{{\mu_x}}}},
\end{equation}

\noindent where $\kappa_{u}$ and $\kappa_{\mu}$ are the gradient descent step sizes. 

\section{Precision Learning for fault-tolerant control}
\label{sec:precision_learning_gradient_on_f}

In previous work \cite{baioumy2021ECC}, the u-AIC is used in combination with an established FT approach to achieve fault detection and recovery. In particular, the sensory prediction errors in the free-energy are used as residual signals for fault detection purposes. The statistical properties of the residuals are analysed offline and healthy boundaries are defined. At runtime, a healthy residual set is computed and if the current residual is outside the admissible set, the relative sensor is marked as faulty. When a fault is detected, the precision (or inverse covariance) of the sensor is abruptly set to zero, that is $P = \Sigma^{-1} = 0$, to exclude that sensor from the optimization of the free-energy. This idea is summarised in Fig.~\ref{fig:FTschemeECC}.

\begin{figure}[htb]
    \centering
    \includegraphics[width=\linewidth]{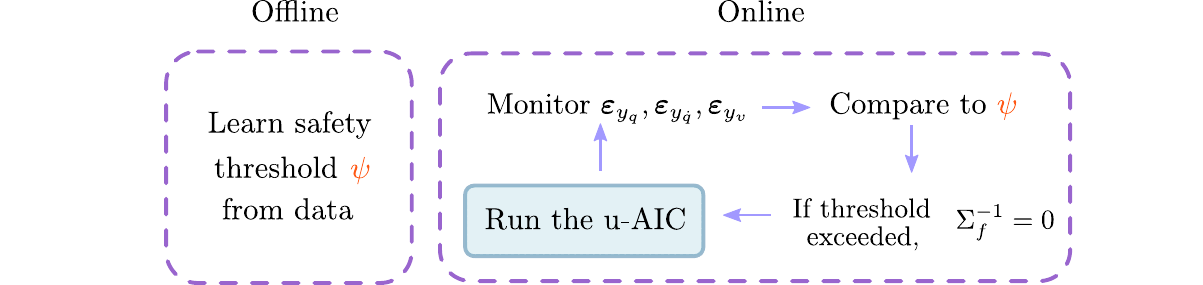}
    \caption{Fault-tolerant pipeline from \cite{baioumy2021ECC}. The term $\Sigma_f^{-1}$ represents the precision of the detected faulty sensor.}
    \label{fig:FTschemeECC}
\end{figure}

In this work, we propose a different approach to achieve fault recovery through online precision learning with u-AIC instead ad-hoc hard switches in the controller's parameters. 
Fig.~\ref{fig:onlineFT} shows thee difference with respect to \cite{baioumy2021ECC}. 

\begin{figure}[htb]
    \centering
    \includegraphics[width=\linewidth]{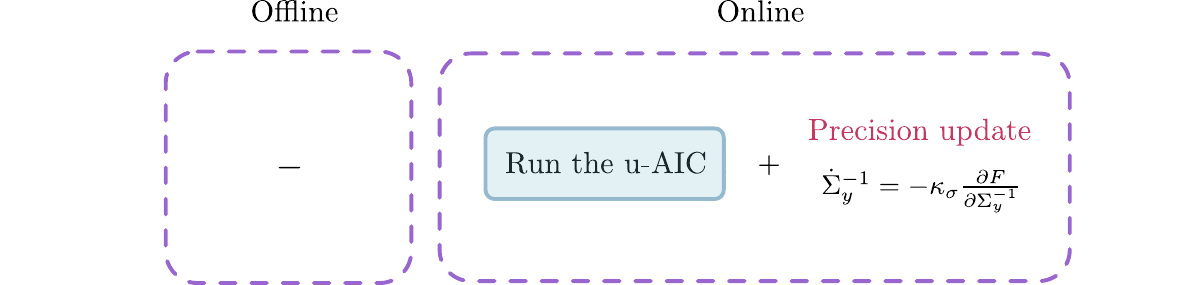}
    \caption{New fault-tolerant pipeline with precision learning, in contrast to previous work \cite{baioumy2021ECC} from Fig.~\ref{fig:FTschemeECC}.}
    \label{fig:onlineFT}
\end{figure}



\subsubsection{Learning sensory precision.}
\label{Learning model variances}

For a sensor $y$, we can update an inverse precision matrix $\Sigma_{y}^{-1}$ using gradient descent on $F$ as done in \cite{2020UKRAS_baioumy,baioumy2020active}:

\begin{equation}
    \label{eq:observaion_variance_update}
    \dot{\Sigma}_{y}^{-1} = -\kappa_{\sigma} \frac{\partial F}{\partial \Sigma_{y}^{-1}}.
\end{equation}

However, we need to ensure that precision remains a positive number. Performing gradient descent does not inherently guarantee that. 


First, consider a one-dimensional problem where state $x$ and observation $y$ are scalars. The observations is affected by zero-mean Gaussian noise with a variance of $\sigma^2$ (also a scalar). The scalar precision is defined as the inverse variance $\omega = 1/\sigma^2$.  As explained, performing gradient descent on the free-energy with respect to $\omega$ may result in it being negative. A simple solution is to perform a reparameterization with a strictly positive function such as an exponential. I.e. we assume that $\omega = \exp{\zeta}$ and we perform gradient descent on $\zeta$: 

\begin{equation}
    \label{eq:exp_variance_update}
    \dot{\zeta} = -\kappa_{\zeta} \frac{\partial F}{\partial \zeta}
\end{equation}

\noindent where $\kappa_{\zeta}$ is the gradient step-size. Another way is to set a lower bound on the variance (as done in \cite{bogacz2017tutorial}). Both methods ensure the variance being positive.

\textbf{{Diagonal precision matrix.}} \\ Guaranteeing a positive semi-definite matrix in an $n$-dimensional case is not as straightforward. However, in the context of a robotic manipulator, one can reasonably assume that the observation noise on each sensor is independent \cite{oliver,Pezzato2020,baioumy2020active}. This means that the covariance (and precision) matrices are diagonal. 

$$
P = \begin{bmatrix}
\omega_1 &  &  & \\ 
 & \omega_2 &  & \\ 
 &  & ... & \\ 
 &  &  & \omega_n
\end{bmatrix}
$$

Given this assumption, every element on the diagonal is positive and can be updated in the same fashion as the scalar case (Eq. \eqref{eq:exp_variance_update}).

\subsubsection{Fault-tolerant control as precision learning.}
Consider the sum of the sensory prediction errors in the free-energy from eq.~\eqref{eq:laplace_F_final_vector}:
\begin{equation}
F  =  \frac{1}{2}(\bm{\varepsilon}_{y_q}^\top \Sigma_{y_q}^{-1}\bm{\varepsilon}_{y_q}
    +  \bm{\varepsilon}_{y_{\dot{q}}}^\top \Sigma_{y_{\dot{q}}}^{-1}\bm{\varepsilon}_{y_{\dot{q}}}
    + \bm{\varepsilon}_{y_v}^\top \Sigma_{y_v}^{-1}\bm{\varepsilon}_{y_v}+...)
    + C,
\end{equation}

Intuitively, when a sensor is faulty, the related sensory prediction error will necessarily be higher since sensory readings and internal beliefs will drift away. After a fault, the estimated precision through our precision learning scheme will be much lower than the original $P = \Sigma^{-1}$. Thus its weight in the free-energy $F$, and so in the state-estimation and control equations as in eq.~\eqref{eq:u-AIC_F_minimize_a} will naturally become lower than the other healthy sensors. Its weight essentially adjusts \textit{proportionally to the degree of the sensor being faulty}. Note that this allows for automatic fault recovery but it does not provide explicit fault detection. In case the latter is needed for a potential user or an additional supervisory system, traditional techniques can be used as the one presented in \cite{baioumy2021ECC} in conjunction with precision learning.

FT control for sensory faults can now be done using precision learning in several ways. The first way is to use it as a stand-alone and activate precision learning for all sensors during operation. In this case, no other methods are needed, no thresholds are designed and the recovery emerges naturally. As mentioned before, the drawback is that the users can not be `alerted' for the presence of a fault (since there is no explicit fault-detection). The second way, which addresses this issue, is to use an established algorithm for fault detection (such as the one presented in \cite{baioumy2021ECC}) and then, only after a fault is detected, allow precision update. 

Interestingly, performing precision learning as presented in this section can make the state-estimation noisier since the agents only relies on the current observation (rather than a batch of last $k$ observations) for the update and both the uncertainty of the state and precision are not quantified. An additional approach would then be to consider the last $k$ observations for the update, but this is out of the scope of this work. 

To summarise, the precision learning in this paper can either be activated at all times or \textit{only after a fault is detected}. Activating the precision learning at all times with a small step-size for the gradient seems to work best.

\section{Results}
We apply the methods in Sec.~\ref{sec:precision_learning_gradient_on_f} on a 2-DOF robotic manipulator. We test three scenarios: a) precision learning at all times, b) precision learning only when a fault is detected and c) a deterministic update as done in \cite{baioumy2021ECC}. Note that the latter has access to a model and uses data to determine a threshold offline. This is not the case for the first two options where only model-free precision learning is performed. The results are summarized in the Table \ref{table: MSE_ft_control}.
\begin{table}
\centering
\begin{tabular}{c|cc}
 &
  \begin{tabular}[c]{@{}c@{}}Joints with\\ encoder fault\end{tabular} &
  \begin{tabular}[c]{@{}c@{}}Joints without\\ encoder fault\end{tabular} \\ \hline
No fault-tolerance   & 0.0036    & 0.0020 \\
PL at all time       & 5.422 e-5 & 4.527 e-5 \\
\begin{tabular}[c]{@{}c@{}}PL + fault-detection \end{tabular} &
  6.097 e-5 &  4.134 e-5 \\
Deterministic fault recovery & \textbf{0.5946 e-5}   & \textbf{0.3579 e-5}
\end{tabular}
\caption{Mean Squared Error (MSE) for different methods of fault-tolerant control. PL indicates precision learning}
\label{table: MSE_ft_control}
\end{table}
In the simulations, the sensors are injected with zero-mean Gaussian noise. The standard deviation of the noise for encoders and velocity sensors is set to $\sigma_q = \sigma_{\dot{q}}  =0.001$, while the one for the camera is set to $\sigma_{v} = 0.01$. The camera is also affected by barrel distortion with coefficients $K_1 = -1.5e^{-3},\ K_2=5e^{-6},$ $K_3=0$ (values are similar to work from \cite{cameraNoise,lipkin}).

The agents starts in configuration $\bm x_0$, then moves to the targets $\bm x_1$ and $\bm x_2$. At $t = 8s$ a fault is injected. The encoder fault is such that the output related to the first joint freezes. For a discrete step $k$ it holds then $\bm y_q(k) = [q_1(k_f),\ q_2(k)]^\top$ for $k \geq k_f$ and $k_f = 8$. The fault detection and recovery of such a fault, as well as the system's response, are reported below in Fig.~\ref{fig:gradient-based-ft_control}.

\begin{figure}[htb]
    \centering
    \includegraphics[width=\linewidth]{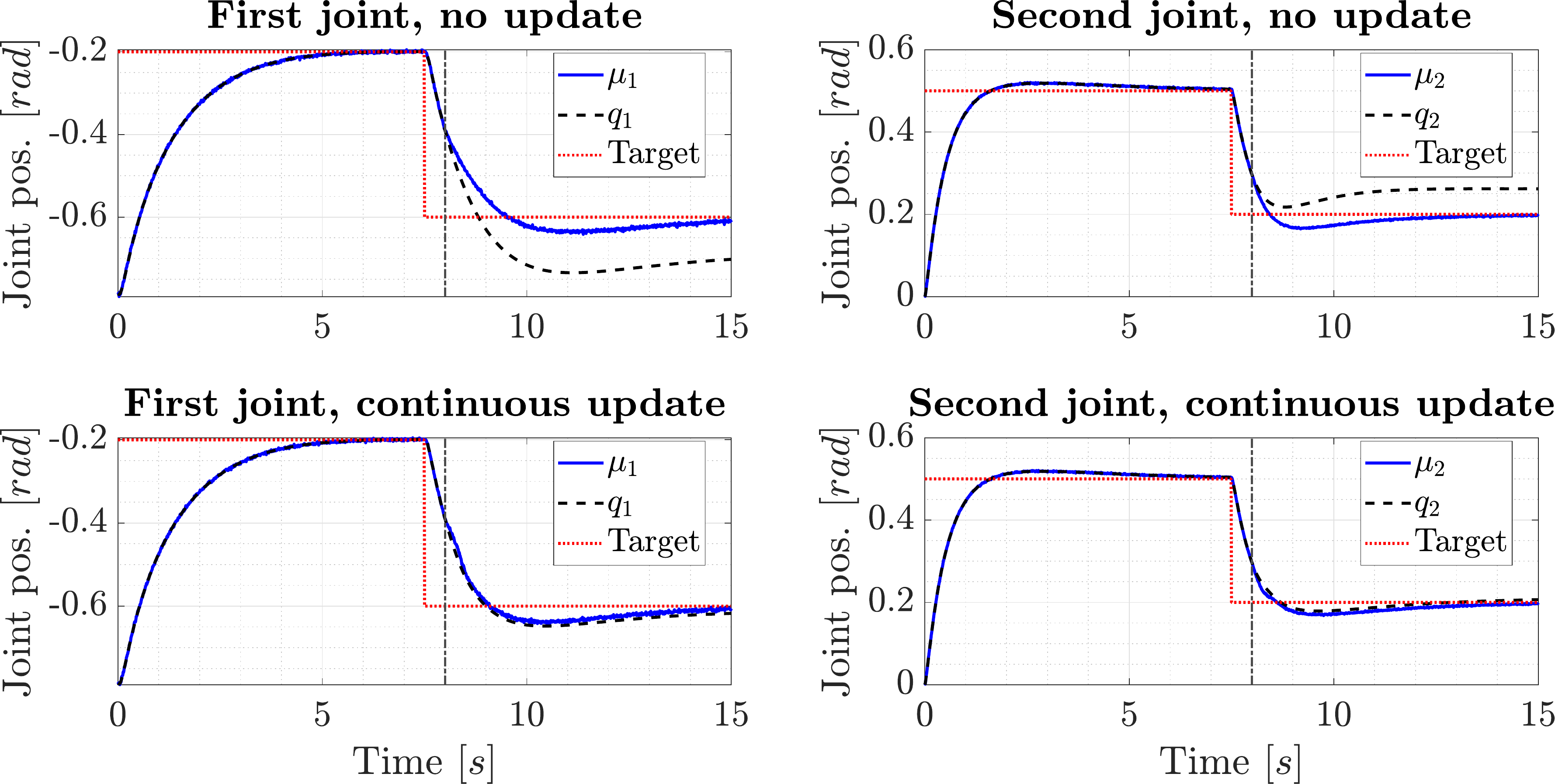}
    \caption{System's response in the case of encoder fault with and without precision learning applied at all times. The fault is injected at $t = 8s$ and indicated with a dot-dashed line.}
    \label{fig:gradient-based-ft_control}
\end{figure}
As seen in Fig.~\ref{fig:gradient-based-ft_control}, the system is not able to reach the set-point after the occurrence of the fault if online precision update is not allowed. The robot arm reaches a different configuration to minimise the free-energy, which is built fusing the sensory information from the (faulty) encoders and the (healthy) camera. However, when the faulty encoder is adjusted using precision learning, the agent is able to reach the final configuration.  

Fig.~\ref{fig:gradient-based-ft_control} reports the results when precision learning is being done during the full operational time. Alternatively, one could only use precision leaning when a fault is detected. This yields a response that is almost identical. The Mean Squared Error (MSE) between the belief and the true position ($\bm \mu_x - \bm x$) is computed on a sample of test runs and reported in the Table~\ref{table: MSE_ft_control}. The results are reported for both the joint whose encoder is faulty, and joints with healthy encoders. In both cases, hard update of the precision to zero has the lowest MSE; however, the approaches based on precision learning do not require any previous information or a threshold definition thus it is simpler to implement. Yet, precision learning has a satisfactory performance while accommodating a sensory fault.

\section{Improving precision learning: a discussion}
\label{sec:discussion}
In this paper, we perform a simple modification to the unbiased active inference controller: adding precision learning for all sensors. We show that this results in stochastic fault-tolerance to sensory faults, i.e. the precision of a faulty sensor will decrease automatically making its relative weight in the control and estimation laws smaller. This eliminates the need to learn a threshold from data offline. Additionally, no ah-hoc recovery action is required. The controller automatically adjusts to the new precision. 

In the experiments, we compared precision learning to a state-of-the-art method. Precision learning was an order of magnitude worse in performance but still satisfactory. Note that precision learning did \textit{not} require any data or training offline to determine thresholds or recovery strategies. Finally, precision learning performs stochastic fault-detection rather than deterministic. 

Most importantly, this approach based on precision learning can be improved in many ways. First, rather than computing a point-mass estimate, we can explicitly model the precision as a random variable and perform inference on it.

We can perform Bayesian inference by modelling the precision as a random variable and computing a posterior over it. In the one dimensional case we use a Gamma prior on the precision $\omega$ as $$\Gamma(\omega ; a, b)  = \frac{b^a}{\Gamma(a)} \omega^{a-1}e^{-\omega b}.$$ Given that the observation model is Gaussian, this choice is beneficial since it is the conjugate prior \cite{bishop2006pattern,murphy2012machine}, where $a$ and $b$ are the parameters of the distribution and $\Gamma(a) = (a - 1)!$  is a factorial function. For example, $\Gamma(5) = 4! = 24$. Now to compute the posterior, we multiply the prior with the Gaussian likelihood model of $p(y|\omega)$ and obtain the posterior which is also a Gamma distribution as shown below.

$$p(\omega) = \Gamma(\omega ; a, b) \propto \omega^{a-1}e^{-\omega b}$$
$$p(\omega|y) \propto p(y|\omega)p(\omega) \propto \omega^{0.5+a-1}e^{-\omega (b + \frac{(y - C)^2}{2})}$$
$$p(\omega|y) = \Gamma(\omega ; a + \frac{1}{2}, b + \frac{(y-C)^2}{2})$$

The last equation shows a simple update rule to modify the belief over the precision for every data point. In the optimization for the state, the following quantities are used: expected precision $\mathbb{E}[\omega] = a/b$, $Mode[\omega] = (a-1)/b$ and $Var[\omega] = a/b^2$. In the $n$-dimensional case, the same procedure can be done but with a Wishart distribution rather than a Gamma.

Additionally, we could use a batch of $k$ observation to learn the precision rather than just one observation. Many approaches for covariance/precision estimation have been successful in robotics e.g. \cite{vega2013cello,pfeifer2017dynamic,vega2013celloEM,shetty2017covariance}. Additionally, many other approaches within the active inference literature can be used for effective precision learning such as dynamic expectation maximization (DEM) \cite{friston2008variational,friston2010generalised}. These will be explored and compared in future work.

\section{Conclusions}
This paper presents a fault-tolerant controller based on active inference. We model the precision (inverse covariance) of each sensor in our system and determine the probability of the sensor being healthy to be proportional to its precision. Rather than reasoning about whether a sensor is faulty or not, we reason about the degree to which the sensor is faulty. We present gradient based approaches to approximate the precision matrices of the system. The results show that the precision learning is a promising approach for fault-tolerant control. It allows for robust behaviour without needing any threshold definition a priori, without designing additional ad-hoc recovery mechanisms, and can be used stand-alone or in conjunction with other methods. The results using precision learning was satisfactory but an order of magnitude away from the to state-of-the-art. However, precision learning was not trained on data offline and performs a stochastic update. Bayesian methods can be used to improve the performance of the approach. Additionally, in all cases regarding precision learning, the performance can be improved by considering the last $k$ observations rather than just one. Future work will address this.

\end{abstract}
%
%


%
%
%
\bibliographystyle{splncs04}
\bibliography{myBib}
%


\end{document}